\documentclass{article}

\usepackage{arxiv}

\usepackage[utf8]{inputenc} 
\usepackage[T1]{fontenc}    
\usepackage{url}            
\usepackage{booktabs}       
\usepackage{amsfonts}       
\usepackage{nicefrac}       
\usepackage{microtype}      
\usepackage{lipsum}
\usepackage{graphicx}

\usepackage{algorithm}
\usepackage{algorithmicx}
\usepackage{algpseudocode}
\usepackage{physics}        
\usepackage{url}            
\usepackage{booktabs}       
\usepackage{nicefrac}       
\usepackage{microtype}      
\usepackage{lipsum}
\usepackage{multirow}
\usepackage{wrapfig}
\usepackage{caption}
\usepackage{subcaption}
\usepackage{xcolor}
\usepackage{todonotes}
\usepackage{xr}
\usepackage{pifont}
\usepackage{makecell}

\DeclareMathOperator*{\argmax}{arg\,max}
\DeclareMathOperator*{\argmin}{arg\,min}

\usepackage[colorlinks=true,urlcolor=blue,citecolor=blue,linkcolor=blue,bookmarks=true]{hyperref}
\usepackage[nameinlink]{cleveref}

\graphicspath{ {./images/} }

\title{Continual Prune-and-Select: Class-incremental learning with specialized subnetworks \thanks{This is a preprint version of the work. The published version is: Dekhovich, A., Tax, D. M., Sluiter, M. H., \& Bessa, M. A. (2023). Continual prune-and-select: class-incremental learning with specialized subnetworks. Applied Intelligence, 1-16. \url{https://doi.org/10.1007/s10489-022-04441-z} }}

\author{
   Aleksandr Dekhovich \\
   Delft University of Technology\\
   \And
   David M.J. Tax\\
   Delft University of Technology\\
   \And
   Marcel H.F. Sluiter\\
   Delft University of Technology\\
   \And
   Miguel A. Bessa \thanks{Corresponding author, e-mail: \texttt{miguel\_bessa@brown.edu}} \\
   Brown University\\
}

\begin{document}
\maketitle
\begin{abstract}
The human brain is capable of learning tasks sequentially mostly without forgetting. However, deep neural networks (DNNs) suffer from catastrophic forgetting when learning one task after another. We address this challenge considering a class-incremental learning scenario where the DNN sees test data without knowing the task from which this data originates. During training, Continual Prune-and-Select (CP\&S) finds a subnetwork within the DNN that is responsible for solving a given task. Then, during inference, CP\&S selects the correct subnetwork to make predictions for that task. A new task is learned by training available neuronal connections of the DNN (previously untrained) to create a new subnetwork by pruning, which can include previously trained connections belonging to other subnetwork(s) because it does not update shared connections. This enables to eliminate catastrophic forgetting by creating specialized regions in the DNN that do not conflict with each other while still allowing knowledge transfer across them. The CP\&S strategy is implemented with different subnetwork selection strategies, revealing superior performance to state-of-the-art continual learning methods tested on various datasets (CIFAR-100, CUB-200-2011, ImageNet-100 and ImageNet-1000). In particular, CP\&S is capable of sequentially learning 10 tasks from ImageNet-1000 keeping an accuracy around 94\% with negligible forgetting, a first-of-its-kind result in class-incremental learning\footnote{Code is available at: \url{https://github.com/adekhovich/continual_prune_and_select}}. To the best of the authors' knowledge, this represents an improvement in accuracy above 10\% when compared to the best alternative method.

\end{abstract}


\section{Introduction}
\label{sec:introduction}

Despite significant progress, deep learning methods tend to forget old tasks while learning new ones. This is known as \textit{catastrophic forgetting} in neural networks \cite{french1999catastrophic,goodfellow2013empirical}. In the conventional setting, a machine learning model has access to the entire training data at any point in time. Instead, in continual learning or lifelong learning \cite{thrun1998lifelong} data for a given task comes in sequentially at a specific learning moment, and then new data associated with another task comes in at a different moment. Continual learning aims at creating deep learning models that do not forget previously learned tasks while being able to learn new ones, i.e. addressing catastrophic forgetting. Classification continual learning problems can be separated into two scenarios \cite{masana2020class}: task-incremental learning (task-IL), where the task being solved is known both during training and inference; and class-incremental learning (class-IL) \cite{rebuffi2017icarl}, where the task-ID is known only during training but unknown during testing. The class-IL problem is notably more challenging than task-IL. There are also incremental learning approaches for object detection \cite{shmelkov2017incremental, zhang2020class} and semantic segmentation \cite{michieli2019incremental, yan2021framework}. However, to the best of our knowledge, there are no examples in the literature of combining different types of learning tasks. Addressing current challenges in incremental learning brings the community one step closer to mimicking the abilities of the human brain. Recently, a brain-inspired replay method \cite{van2020brain} was proposed to generate an internal feature representation in order to reduce forgetting. The idea behind this technique is to be able to generate features of old data, imitating memory mechanisms in the brain.

There is evidence from neuroscience \cite{lerner2011topographic, zadbood2017we,huttenlocher1990morphometric} that humans have special regions in the brain that are responsible for the recognition of specific patterns. Moreover, several studies show that the human brain encodes information in a sparse representation with an optimal fraction of active neurons of 1\%-4\% at the same time \cite{lennie2003cost, attwell2001energy}. Motivated by this observation, we propose a class-IL algorithm for image classification based on two steps: creating a subnetwork for a given task during training and selecting a previously obtained subnetwork during inference to make predictions. The first stage is achieved via iterative pruning that propagates input patterns through the network and eliminates the least useful connections. During inference, we first predict the current task when selecting the appropriate subnetwork from a small batch of test samples, and only then make a prediction with the selected subnetwork. We allow overlaps between subnetworks in order to induce knowledge transfer during training of new tasks. However, previously trained weights are not changed. Parameter update only occurs when training available neuron connections, which become part of a new subnetwork associated with the new task. 

\paragraph{Our contribution} To the best of our knowledge, we propose for the first time a general strategy to create overlapping subnetworks of neuronal connections that share knowledge with each other for the class-IL scenario. In doing so, we achieve a strategy with negligible forgetting, unlike other works to date. Importantly, the choice of methods for subnetwork creation and for subnetwork selection \textit{can be} different from the ones considered herein. Our goal is to propose a simple working paradigm for class-IL problems, where firstly some connections are assigned to a specific task during training that can then be selected during inference \textit{without} knowing the task-ID.

This paper starts discussing state-of-the-art methods for continual learning, their differences and fundamental assumptions (Section \ref{sec:related_work}). Then, the proposed CP\&S strategy is presented (Section \ref{sec:proposed_method}), and evaluated (Section \ref{sec:experiments}) on class-incremental learning scenarios constructed from various datasets, including CIFAR-100 \cite{krizhevsky2009learning}, ImageNet-100, ImageNet-1000 \cite{deng2009imagenet}, and CUB-200-2011 \cite{WahCUB_200_2011}. In Section \ref{sec:further_analysis}, additional experiments are provided showing limitations of CP\&S approach and opening avenues for the conclusions and future directions (Section \ref{sec:conclusion}).

\section{Related Work}
\label{sec:related_work}

\paragraph{Class-incremental learning.} As previously mentioned, continual learning problems are usually classified according to whether or not the task-ID is available during inference. We focus on the class-IL scenario where the task ID is absent during inference since it is the most realistic and challenging scenario of continual learning. All class-IL methods are usually divided into three categories \cite{delange2021continual}: \textit{regularization} \cite{zenke2017continual,li2017learning,dhar2019learning,liu2018rotate,kirkpatrick2017overcoming,aljundi2018memory}, \textit{rehearsal} \cite{rebuffi2017icarl,hou2019learning,belouadah2019il2m,douillard2020podnet} and \textit{architectural} \cite{yoon2017lifelong,wortsman2020supermasks,sokar2021spacenet}.

Purely \textit{regularization-based} methods introduce an additional term in the loss function to prevent forgetting. Some approaches \cite{kirkpatrick2017overcoming,zenke2017continual,aljundi2018memory,chaudhry2018riemannian} estimate the importance of connections for a given task and penalize the model for gradient updates during training for the next tasks. Learning without forgetting (LwF) \cite{li2017learning} adds a term in the loss function that penalizes changes in old output heads for new data while training new output heads on new data. Regularization-based methods have the advantage of not storing past data in memory nor needing network expansion, but they perform worse compared to other approaches \cite{masana2020class}.

\textit{Rehearsal} methods replay small amounts of old classes \cite{rebuffi2017icarl} or generate synthetic examples \cite{shin2017continual} to be able to predict previously seen classes. iCaRL uses the nearest class mean \cite{mensink2012metric} (NCM) classifier together with fixed memory of old data to mitigate forgetting. Bias-correction methods \cite{castro2018end,hou2019learning,wu2019large} aim to tackle the tendency of class-IL algorithms to be biased towards classes of the last tasks, which arises due to class imbalance at the latest stages \cite{masana2020class}. PODNet \cite{douillard2020podnet} has multiple terms in the loss function using old data from the memory of a fixed size, penalizing signal deviations not only in the output layer but also in intermediate ones. AFC \cite{kang2022class} uses knowledge distillation by estimating the importance of each feature map. The estimation is based on the increase of loss function from changing channels' parameters in the feature maps. The obvious limitation of rehearsal methods is the need to keep past data, which is often not desirable in practical applications due to privacy issues as mentioned by \cite{zhang2020class}. 

\textit{Architectural} methods follow a different strategy where the network architecture is modified to avoid forgetting. For example, the Dynamically Expandable Network (DEN) \cite{yoon2017lifelong} expands the network architecture in an online manner, increasing network capabilities, and introducing a regularization term to prevent forgetting. However, due to the expansion of the network, the final number of parameters is greater than for the original architecture, which increases the memory costs. Supermasks in Superposition (SupSup) \cite{wortsman2020supermasks} and SpaceNet \cite{sokar2021spacenet} find subnetworks for every task. SpaceNet assigns parameters to one task only, without sharing knowledge between tasks which limits its allocation capabilities for long sequences of tasks. In addition, SpaceNet requires to pre-define the sparsity level for each task. SupSup uses a randomly weighted backbone \cite{rajasegaran2019random} instead of pruning to obtain a task-related subnetwork. During inference, SupSup predicts the correct subnetwork for the given test data, using all data points in the batch. In the provided experiments, the batch size is equal to 128 images which may not be applicable to real-life problems. DER \cite{yan2021dynamically} dynamically expands the feature extractor by introducing new channels and freezes old feature representation while learning a new task. Later, DER uses a small portion of old data and current data to finetune the network for all tasks. To stop the growth of the number of parameters, DER uses a pruning strategy, however, the final number of parameters is unpredictable. Similarly, FOSTER \cite{Wang2022FOSTERFB} introduces a new module with a feature map to learn new classes. However, it uses a knowledge distillation strategy inspired by gradient boosting instead of pruning to compress the model. As a result, the outcome of FOSTER is a single fixed-sized backbone network.

A Meta-Learning approach for class-IL is proposed by iTAML \cite{rajasegaran2020itaml}. The algorithm for updating parameters for all old tasks also needs fixed-sized memory, but iTAML uses a momentum-based strategy for meta-updates to overcome catastrophic forgetting. At the test stage, iTAML starts by predicting the task associated with that sample using a given test batch, and then adapts its parameters to the predicted task using data from fixed memory. Finally, with the adapted model and predicted task-ID, iTAML makes a prediction. Overall, iTAML uses samples from previous tasks to prevent forgetting, and the batch of test data to predict task-ID, making it the most demanding algorithm out of consideration. Also, the model adaptation to the predicted task makes it computationally more expensive than other state-of-the-art methods.

\paragraph{Iterative pruning for Continual learning.} Typically, neural network pruning is used for model compression such that it reduces memory and computational costs. The pruning pipeline consists of three steps: network pretraining, deleting the least important connections or neurons based on some criterion, and network retraining. Iterative pruning is characterized by repeating the second and third steps several times. There are numerous approaches to pruning, namely magnitude pruning \cite{han2015learning,li2016pruning,frankle2018lottery}, data-driven pruning \cite{hu2016network,huang2018data,luo2017thinet,dekhovich2021neural} and sensitivity-based pruning \cite{lecun1990optimal,hassibi1993optimal,hassibi1993second,lebedev2016fast}. Iterative pruning has been recently applied in the context of task-IL but not in the more challenging class-IL scenario, where the task-ID is not known \textit{a priori}. Unsurprisingly, pruning has been shown to lead to simplified neural networks with a small fraction of the original parameters. This can facilitate the accumulation of knowledge for new tasks, as demonstrated by task-IL methods based on iterative pruning, namely PackNet \cite{mallya2018packnet} that uses magnitude connections pruning \cite{han2015learning}, and CLNP \cite{golkar2019continual} that uses data-driven neurons pruning \cite{huang2018data}. Piggyback \cite{mallya2018piggyback} learns the mask for every task, as well as CPG \cite{hung2019compacting} which also expands a network in the ProgressiveNet manner \cite{rusu2016progressive}. The performance of these algorithms is strong for task-IL, but they have the significant limitation of requiring to know the task-ID.

We are interested in developing a class-IL method that contains the benefits of iterative pruning connections to obtain sparse network representations while being capable of selecting tasks without knowing the task-ID. To this effect, we developed a pruning strategy called NNrelief \cite{dekhovich2021neural} that aims at leaving as many connections as possible available for future tasks, leading to sparser networks when compared to other pruning methods. The algorithm's idea is to propagate signal through the network, compute a metric called importance score for each connection which estimates its contribution to the signal of the following neuron, and then prune the least contributing connections incoming to the neuron.

\paragraph{Task selection.} Currently, there are few strategies for task selection in class-incremental learning. For example, iTAML \cite{rajasegaran2020itaml} and SupSup \cite{wortsman2020supermasks} use similar ideas for task identification class-IL applied to image classification problems, and neither method uses pruning as a means to create space for new knowledge. The underlying assumption is that if a classifier network is well-trained, the highest output signal in the neuron of the output layer corresponds to the class belonging to the correct task. So, iTAML sums the largest output values of every task-related output in that layer over every test image in the batch, and then finds the layer with the highest total sum. SupSup relies on the entropy of the signal in each of the heads, in the hope that the model is confident in its prediction when it is in the correct head, meaning that the entropy of the signal within the head should be smaller than in other heads. Note that both methods use batches of test samples to select the correct task: iTAML varies the batch size from 20 to 150 depending on the dataset, while SupSup uses 128 images in their experiments. A different strategy is pursued by Kim et al. \cite{kim2020class}, where an autoencoder is associated with a task during training. In the test stage, the reconstruction loss is computed for \textit{every} autoencoder with the given test image, and the one with minimum reconstruction loss is chosen to make predictions. Subsequently, a classification model makes a prediction with the given predicted task-ID. It was shown that in the case of LwF \cite{li2017learning} and LwM \cite{dhar2019learning} this task-selection procedure improves classification accuracy. However, this task-selection approach requires training an autoencoder for every task which is impractical.

\paragraph{Limitations of class-IL approaches.} State-of-the-art class-IL methods have simplified training by replaying old data \cite{rebuffi2017icarl,douillard2020podnet,rajasegaran2019random,hou2019learning}, doing inference with a batch of images to determine the current task \cite{rajasegaran2020itaml,wortsman2020supermasks} and by performing adaptation before inference \cite{rajasegaran2020itaml}. Table \ref{tab:assumptions} summarizes these assumptions for each method. In rehearsal methods, examples of previous classes are stored (with fixed or growing memory), which makes them inappropriate when images should not be kept for a long time. Similarly, the adaptation of a model for a given batch of test data before making a prediction (as in iTAML) is only possible when having examples in memory. Furthermore, the need for a significant number of images in a batch during inference arises from the difficulty of identifying the task-ID correctly with one image only. These can be strong model constraints when considering real-life applications.

\begin{table}
\begin{center}
    \begin{minipage}{\linewidth}
    \caption{Assumptions used by different types of class-IL methods (``bs'' means batch size).}
    \small
    \centering
    \begin{tabular}{@{}l p{0.5in} p{0.35in} c@{}}
        \toprule
        Methods & Replay old data & test $bs > 1$ & Adaptation \\ 
        \midrule
        SI, MAS, LwF, LwM, SpaceNet  & \textcolor{teal}{no} & \textcolor{teal}{no} &  \textcolor{teal}{no} \\ 
        iCaRL, BiC, PODNet, DER, AFC, FOSTER &  \textcolor{red}{yes} &  \textcolor{teal}{no} &  \textcolor{teal}{no} \\
        iTAML & \textcolor{red}{yes} &  \textcolor{red}{yes} &  \textcolor{red}{yes} \\
        \midrule
        \textbf{CP\&S} (ours) & \textcolor{teal}{no} & \textcolor{red}{yes} & \textcolor{teal}{no} \\ 
        \bottomrule
    \end{tabular}
    \label{tab:assumptions}
    \end{minipage}
\end{center}    
\end{table}

\label{sec:proposed_method}

\begin{figure}[ht]
    \centering
    \includegraphics[width=\textwidth]{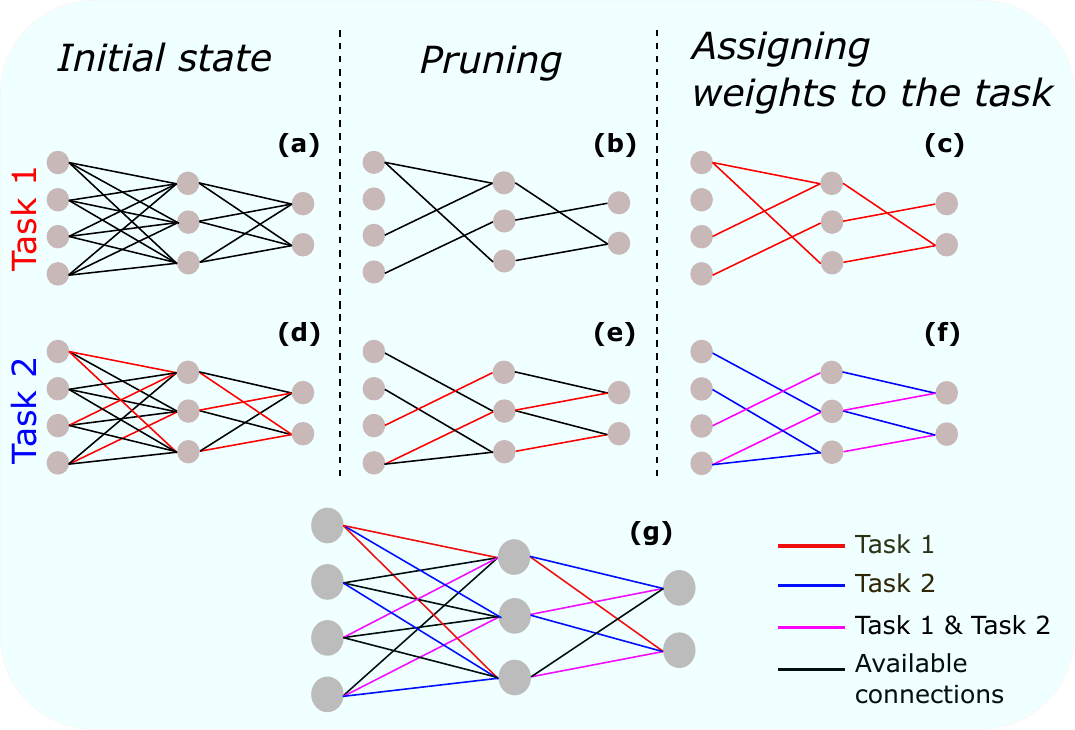}
    \caption{The overview of Continual Prune-and-Select (CP\&S) training procedure. In stages (a)--(c), the first task is learned; in (d)--(f) the network learns task 2; the final outcome is in (g), where the network is trained for both tasks.} \label{fig:cps_pipeline}%
\end{figure}

Our method (CP\&S) is based on training a subnetwork for each given task and then selecting the correct subnetwork when doing inference for new data with an unknown task-ID. We start by training a regular neural network for a specific task (Figure \ref{fig:cps_pipeline}(a)), iteratively pruning it to find a subnetwork with good performance (Figure \ref{fig:cps_pipeline}(b)). This creates a trained subnetwork capable of performing that particular task, leaving the remaining network free for future tasks (Figure \ref{fig:cps_pipeline}(c)). Importantly, when a new task comes in (Figure \ref{fig:cps_pipeline}(d)), the corresponding new subnetwork is found by iteratively pruning the entire original network -- including all free neuronal connections \textit{and} all existing subnetworks found for other tasks (Figure \ref{fig:cps_pipeline}(e)). This is possible by freezing the parameters of previously found subnetworks (avoiding to forget past tasks associated with the corresponding subnetworks), updating all the remaining parameters of the network, and then pruning the entire network until the corresponding subnetwork is found. This way, the new subnetwork (Figure \ref{fig:cps_pipeline}(f)) can contain connections from other subnetworks but it does not affect their performance on past tasks because it did not update the parameters of \textit{shared connections} -- it only updated the parameters of \textit{unshared connections}. This allows to have the transfer of knowledge from one task to another without forgetting (Figure \ref{fig:cps_pipeline}(g)).

The new strategy proposed herein can be implemented with different pruning algorithms to create each subnetwork, and with different task selection algorithms to find the correct subnetwork for inference. As long as the pruning and selection strategies have reasonable performance, we expect this strategy to outperform previous continual learning methods in the class-IL scenario because (1) it avoids forgetting if the task is selected correctly (unlike iTAML); (2) it allows knowledge transfer among tasks (unlike SpaceNet). Also, CP\&S (3) does not need to replay old data (unlike iCaRL, RPS-net, BiC, LUCIR, PODNet, AFC); (4) The backbone architecture is fixed and never changes during training (unlike DER), or requires additional temporary modules (unlike FOSTER).

Without loss of generality, we use our recently developed NNrelief \cite{dekhovich2021neural} pruning algorithm because it promotes sparser networks when compared to the state of the art, and it creates a renormalization effect in the network that distributes the importance of neuronal connections. Concerning task selection (in our case subnetwork selection), we considered different strategies, including the one proposed in the literature that applies to our method (see iTAML \cite{rajasegaran2020itaml} and SupSup \cite{wortsman2020supermasks}). 

Formally, denoting our classification network as $\mathcal{N}$ and considering $T$ tasks, then: 
\begin{equation}
    \mathcal{N} = \cup_{t=1}^{T} \mathcal{N}^t,  
\end{equation}
where $\mathcal{N}^t$ is the subnetwork for task $t$, with $t=1, 2, \ldots, T$. Each subnetwork $\mathcal{N}^t$ is found with our NNrelief pruning algorithm that determines the most important parts of the main network for solving a given task $t$. 
This algorithm estimates each connection's contribution to the total signal of a receiving neuron when compared to the other connections that are incoming to that neuron. This contribution is computed by the \textit{importance score} (IS) of every connection:

\begin{equation}
    \label{eq:fc_IS}
    s_{ij} (\mathbf{x}_1, \mathbf{x}_2, \ldots, \mathbf{x}_N) = \frac{\overline{\lvert w_{ij} x_{i}\rvert}}{\sum_{k=1}^m \overline{\lvert w_{kj} x_{k}\rvert} + \lvert b_j \rvert},
\end{equation}

\noindent for the input signal $\mathbf{X} = \{\mathbf{x}_1, \ldots, \mathbf{x}_N\}$ with $N$ data points, $\mathbf{x}_n = (x_{n1}, \ldots, x_{nm}) \in \mathbb{R}^{m}$ for $m$ neurons in that layer, where $\overline{\lvert w_{ij} x_{i}\rvert} = \frac{1}{N}\sum_{n=1}^N \lvert w_{ij}x_{ni}\rvert$ and with $w_{ij}$ being the weight of the connection between neurons $i$ and $j$, where $b_j$ is the bias in neuron $j$. Then NNrelief prunes the connections entering the neuron with the lowest contribution to the importance score whose sum is less than $(1-\alpha)\sum_{i=1}^m s_{ij}$, where $\alpha$ is the hyperparameter of the algorithm, $\ 0 \le \alpha \le 1$. More details are given in our original article \cite{dekhovich2021neural}.

\begin{algorithm}[H]
    \begin{algorithmic}[1]
        \Require network $\mathcal{N}$, datasets $\{\textbf{X}^t\}_{t=1}^T$.
        Initialize {learning parameters $p$ (learning rate, weight decay, number of epochs, etc. ), pruning parameters (for NNrelief algorithm: $\alpha$ and the number of pruning iterations $k$)}
        \For {$t=1, 2, \ldots, T$}
            \State{$\mathcal{N}^t \gets \text{Pruning}(\mathcal{N}, \textbf{X}^t, \alpha, k)$}
            \State{freeze parameters $w \in \mathcal{N}^t$ and never update them}
        \EndFor
        \Ensure network $\mathcal{N}$ that learned tasks $1, 2, \ldots, t+1$.
    \end{algorithmic}
    \caption{Pseudocode for CP\&S training procedure}
    \label{alg:cps_training}
\end{algorithm}

In the context of class-IL we receive datasets $\textbf{X}^1, \textbf{X}^2, \ldots, \textbf{X}^T$ sequentially. The pruning algorithm then creates masks $\textbf{M}^1, \textbf{M}^2, \ldots, \textbf{M}^T$ for every task $t = 1, 2, \ldots, T$, where $\textbf{M}^t = (m^t_{ij})_{i,j}, \\ m^t_{ij} = 
\begin{cases}
    1, & \text{if there is an active connection} \\ & \text{between neurons $i$ and $j$,}\\
    0, & \text{ otherwise}
\end{cases}$

\noindent and corresponding importance scores $S^1, S^2, \ldots, S^T, \ S^t = (s^t_{ij})_{i,j} $. 

Once the subnetworks are created during training, selecting the correct subnetwork given a batch of test data becomes essential to do inference. In this article, we define a test batch of size $s$ as $\mathbf{X}^{test} = \{\mathbf{x}^{test}_1, \mathbf{x}^{test}_2, \dots, \mathbf{x}^{test}_s\}$, and can simplify the notation to cases where the fully connected part of the network consists of one layer since we run all our experiments on ResNet architectures. However, there are no restrictions to apply this approach to any other type of architecture. We define the convolutional part for task $t$ as $\theta^t, \ \theta^t : \mathbb{R}^{3\times H \times W} \to \mathbb{R}^d$ ($H, W$ are the height and width of an input image and $d$ is the length of an output feature vector), and the fully connected layers as $\varphi^t : \mathbb{R}^d \to \mathbb{R}^{num\_classes}$. 

Similarly to the selection of the pruning algorithm for subnetwork creation, we can also adopt different strategies to identify the correct subnetwork associated with a particular task. In order to establish a fair comparison with the literature, we focus on the \textit{maximum output response} (maxoutput) strategy that is used by other methods (e.g. \cite{rajasegaran2020itaml,wortsman2020supermasks}), but we also show that other task selection methods can lead to good results (see Appendix for a strategy based on Importance Scores). 

The maxoutput strategy for task prediction is simply formulated as:
    \begin{equation}\label{eq:maxoutput}
        t^{*} = \argmax_{t=1,2, \ldots, T}{\sum_{i=1}^{s}{\max \varphi^t(\theta^t(\mathbf{x}^{test}_i))}}.
    \end{equation}
    
This does not require data storage. There are no memory costs associated with this prediction.

\begin{algorithm}[H]
    \begin{algorithmic}[1]
        \Require network $\mathcal{N}$, test batch $\mathbf{X}^{test}$.
        \State predict task $t^{*}$ for the test batch $\mathbf{X}^{test}$ using Eq. \eqref{eq:maxoutput}
        \State make a prediction $\mathbf{\hat{y}} = \mathcal{N}^{t^{*}}(\mathbf{X}^{test})$
        \Ensure predicted classes $\mathbf{\hat{y}}$ for test data $\mathbf{X}^{test}$.
    \end{algorithmic}
    \caption{Pseudocode for CP\&S inference procedure}
    \label{alg:cps_inference}
\end{algorithm}

\section{Experiments}
\label{sec:experiments}

We compare CP\&S with different methods available in the literature. Aiming to establish a fair comparison, we use different measurements of accuracy during the learning process, namely \textit{average multi-class accuracy} (ACC), \textit{backward transfer metric} (BWT) \cite{lopez2017gradient} and \textit{average incremental accuracy} (AIA) \cite{rebuffi2017icarl}. These metrics can be written assuming that a model learned $T$ tasks and denoting $R_{t_2, t_1}$ as the accuracy for task $t_1$ after learning up to task $t_2$ (inclusive, i.e. $t_2 \ge t_1$):
\begin{align}
    \text{ACC}(T) &= \frac{1}{T} \sum_{t=1}^T R_{T, t} \\
    \text{BWT}(T) &= \frac{1}{T-1} \sum_{t=1}^{T-1} R_{t,t}-R_{T,t} \\
    \text{AIA}(T) &= \frac{1}{T} \sum_{t=1}^{T} \text{ACC}(t) 
\end{align}

The idea of the BWT is to measure the forgetting of the incremental-learning models, evaluating how much information about previous tasks is lost after learning a new one. We evaluate all methods using several class orderings to obtain robust results, as recommended in \cite{masana2020class}.

\paragraph{Datasets.} We evaluate CP\&S on three datasets: ImageNet-1000, including its subset ImageNet-100 \cite{deng2009imagenet}; CUB-200-2011 \cite{WahCUB_200_2011}; and CIFAR-100 \cite{krizhevsky2009learning}. We also consider different task construction scenarios. For completeness, the datasets are briefly described as follows:
\begin{itemize}
    \item ImageNet-1000 consists of 1,281,167 $224\times224$ RGB images for training and 50,000 images for validation of 1000 classes. We split both ImageNet-100 and ImageNet-1000 in 10 incremental steps of equal size, similarly to the literature;
    \item CUB-200-2011 consists of 11,788 $224\times224$ RGB images of 200 classes with 5,994  training and 5,794 for testing images;
    \item CIFAR-100 consists of 60,000 $32\times32$ RGB images of 100 classes with 6k images per class. There are 50,000 training samples and 10,000 test samples;
\end{itemize}

We start our experiments with ImageNet-100 (the first 100 classes of the ImageNet-1000 dataset) and with CIFAR-100 before considering more challenging datasets such as ImageNet-1000 and CUB-200-2011. For all datasets, we use the ResNet-18 architecture, as considered by previous methods. For ImageNet-100/1000 datasets, we split them into 10 tasks of the same size (each task having 10 classes). We compare CP\&S with other state-of-the-art models, namely iCaRL \cite{rebuffi2017icarl}, EEIL \cite{castro2018end}, BiC \cite{hou2019learning}, RPS-net \cite{rajasegaran2019random}, iTAML \cite{rajasegaran2020itaml}, DER \cite{yan2021dynamically} and FOSTER \cite{Wang2022FOSTERFB}. In addition, we provide a comparison with the case of \text{Finetuning}, when no anti-forgetting actions are performed, and a network sequentially learns new tasks one by one. For comparison with other works, we either reproduce the results from the official GitHub repository using the hyperparameters mentioned in the original articles or report the results from the original works when available. See the details in Appendix \ref{sec:cifar100_appendix}.

\begin{figure}
    \centering
    \includegraphics[width=\textwidth]{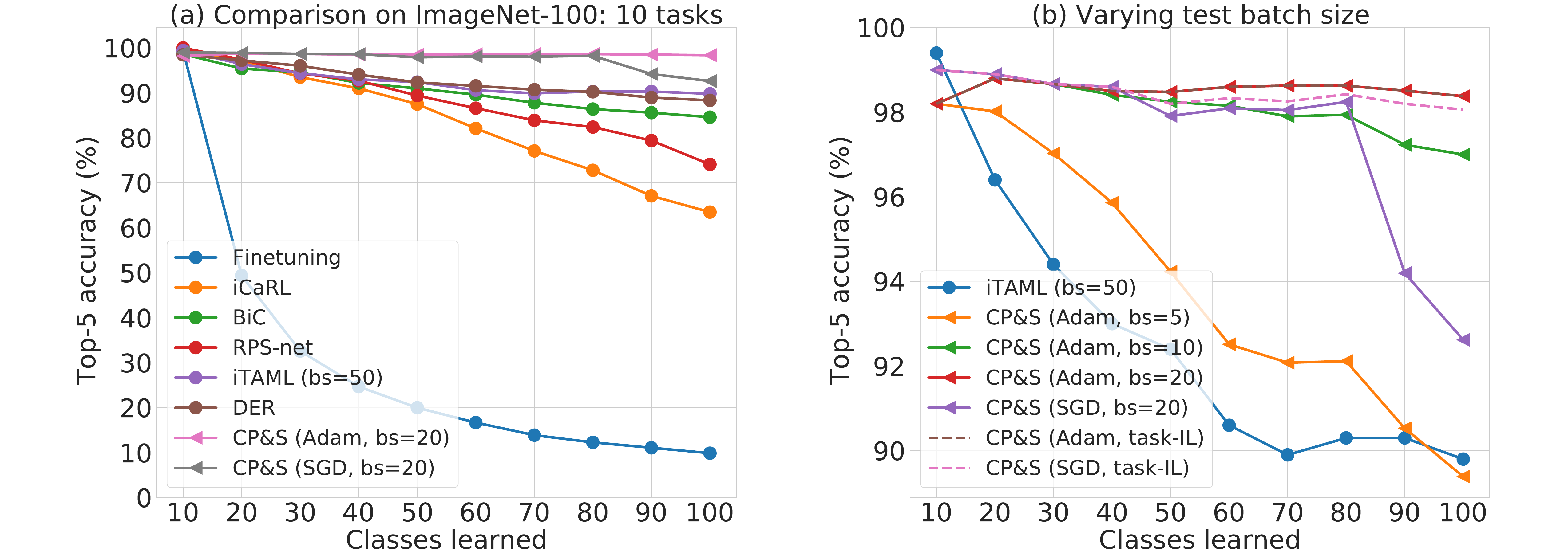}
    \caption{Results on ImageNet-100 and comparison with other approaches. Notation: "bs" refers to the test batch size; "task-IL" refers to the task-IL scenario where the task-ID is known, providing an upper bound to the results. The pruning parameter of CP\&S is $\alpha_{conv}=0.9$ for both optimizers, SGD and Adam. The class ordering is generated by seed 1993 (iCaRL seed).} \label{fig:imagenet100_resnet18_acc}
\end{figure}

\paragraph{ImageNet-100.} Figure \ref{fig:imagenet100_resnet18_acc} shows that CP\&S outperforms state-of-the-art methods for this dataset, even when considering two different optimizers -- Stochastic Gradient Descent (SGD) and Adam \cite{kingma2014adam}. We use a learning rate of $0.1$ for SGD and $0.01$ for Adam, dividing it by 10 on epochs 30 and 60, and we consider weight decay of $10^{-4}$ for both optimizers. Figure \ref{fig:imagenet100_resnet18_acc}(a) shows our predictions using a smaller batch size than iTAML -- 20 samples instead of 50 -- and compares them with other methods. Figure \ref{fig:imagenet100_resnet18_acc}(b) clarifies the influence of considering different test batch sizes in CP\&S method, where it is demonstrated that even when using 5 or 10 samples per batch we still perform better. The same figure also shows that when using Adam we identify the correct subnetwork in $100\%$ of the cases because we reach the upper bound provided in the task-IL scenario, i.e. where the task-ID is known and subnetwork selection is not necessary. For SGD, we observe a slight drop in accuracy after task 8 compared to the task-IL scenario, although it  still outperforms iTAML even though the latter uses 50 images for task identification and requires keeping images in memory. Overall, CP\&S reaches $98.38\%$ accuracy with Adam and $92.62\%$ with SGD, translating into improvements for this dataset beyond $8\%$ and $2.5\%$ when compared to next best method, and even larger when compared to other methods after all classes are learned.

We note that using Adam \cite{kingma2014adam} is advantageous for CP\&S due to the higher level of sparsity that is produced after pruning with NNrelief when compared with other optimizers \cite{dekhovich2021neural}. Note that pruning makes neuron connections available for creating new subnetworks associated with future tasks. If the number of available connections is small, then new subnetworks may not be sufficiently expressive to reach high accuracy for a given task. A similar effect is expected if the number of tasks is large, as shown in the next experiments for CIFAR-100.

\paragraph{CIFAR-100.} Before considering more challenging datasets such as ImageNet-1000 and CUB-200-2011, we focus on CIFAR-100 where we split its 100 classes by a different number of tasks: 5, 10 and 20 tasks composed of 20, 10 and 5 classes, respectively. For iTAML, we followed the original implementation with hyperparameters described in the paper including the test batch size equal to 20, and using \textit{ResNet-18(1/3)} \cite{lopez2017gradient} which is a modified version of the standard ResNet-18 architecture where the number of filters is divided by three. For CP\&S, we use Adam for training using 70 epochs and starting with the learning rate $0.01$ which is then divided by 5 every 20 epochs. We use $\alpha_{conv}=0.9$  and 3 pruning iterations as the pruning parameters of NNrelief.

\begin{figure} 
  \centering
  \includegraphics[width=\linewidth]{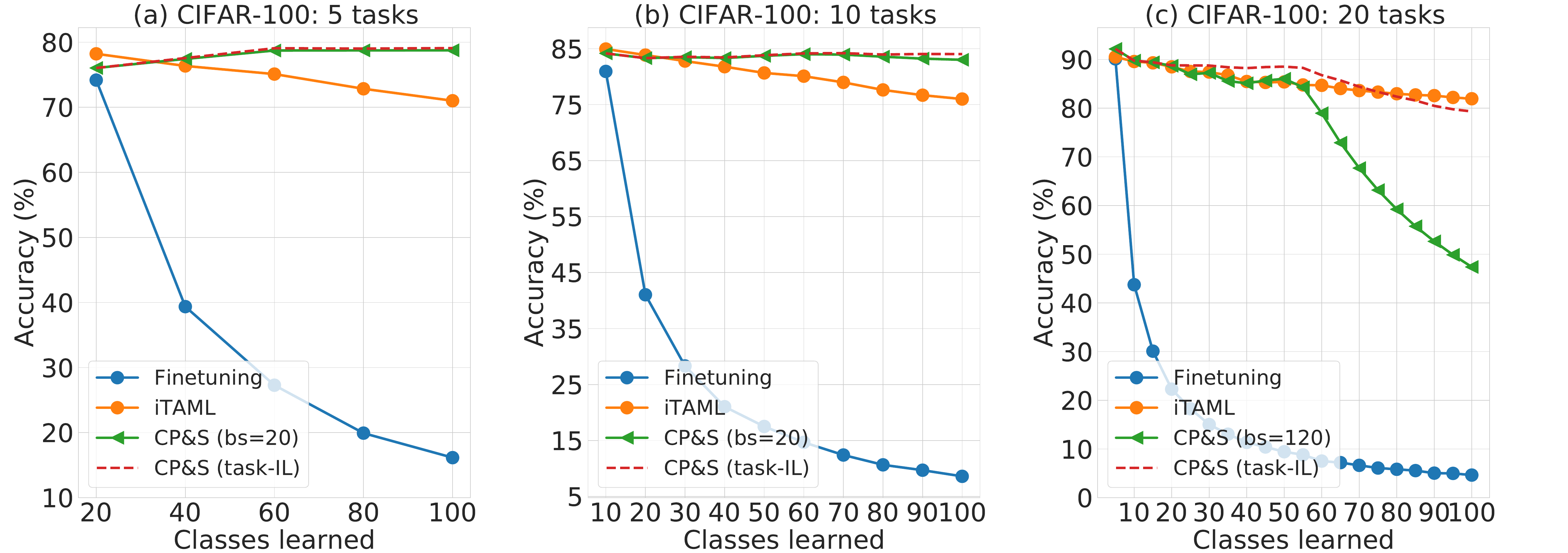}
  \caption{Comparison with iTAML on CIFAR-100 split in 5, 10 and 20 tasks. Notation: ``bs'' refers to the batch size during inference, ``task-IL'' refers to the task-IL scenario where the task ID is known, providing an upper bound to the results. Five different class orderings are used.} \label{fig:cifar_100_itaml}
\end{figure}

\begin{figure}[ht]
  \centering
  \includegraphics[width=\textwidth]{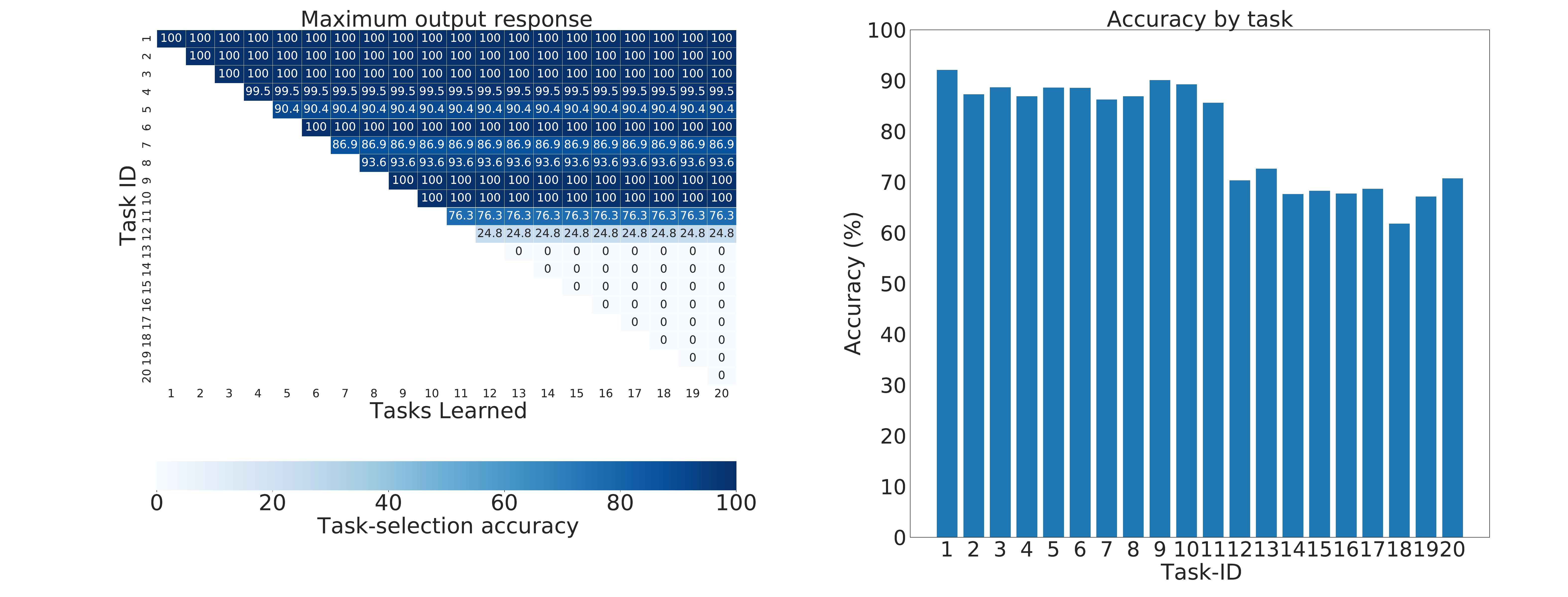}
  \caption{CIFAR-100 divided into 20 tasks of 5 classes each. Task-selection accuracy with maxoutput (\textbf{left}) and accuracy by task (\textbf{right}). } \label{fig:cifar_100_20tasks}
\end{figure}

Figure \ref{fig:cifar_100_itaml} shows that when considering 5 or 10 tasks  CP\&S significantly outperforms iTAML with the same batch size. However, for twenty tasks our performance drops sharply after the eleventh task, even in the ideal case where the task-ID is given (task-IL scenario). Despite being able to alleviate this drop by considering a larger test batch size, or by considering a different strategy for task (subnetwork) selection based on Importance Scores (see Appendix \ref{sec:cifar100_appendix}), we observe this drop occurs approximately at the same number of tasks, independently of the class ordering used. 
Figure \ref{fig:cifar_100_20tasks} (left) provides an explanation of what occurs for the 20 tasks case by showing a heatmap with the task-selection accuracy by row for every task after a new task is learned. We also evaluate the prediction accuracy for each task when the task-ID is known (task-IL scenario) so that we can isolate the effect of not being able to appropriately select the subnetwork of interest for a given task and the effect of achieving low accuracy for a specific task. We observe that even in this case, the accuracy for each new task after the eleventh also drops (see Figure \ref{fig:cifar_100_20tasks} (right)). Therefore, our method performs well until we reach a saturation point when there are not enough neuron connections available to create a sufficiently large subnetwork to achieve high prediction accuracy for a new task. This is a logical conclusion, as one can only learn new tasks while sufficient neuronal connections remain available for training. Increasing the size of the original architecture eliminates this issue.

In addition, note that we do not keep data in memory (no replay), nor do we need to use adaptation to estimate the task before making a final prediction, unlike the methods reviewed above. We also use smaller test batch sizes than iTAML, despite using the same task selection strategy. The following experiments show that this conclusion holds for more challenging datasets.

\begin{figure}
  \centering
  \includegraphics[width=\linewidth]{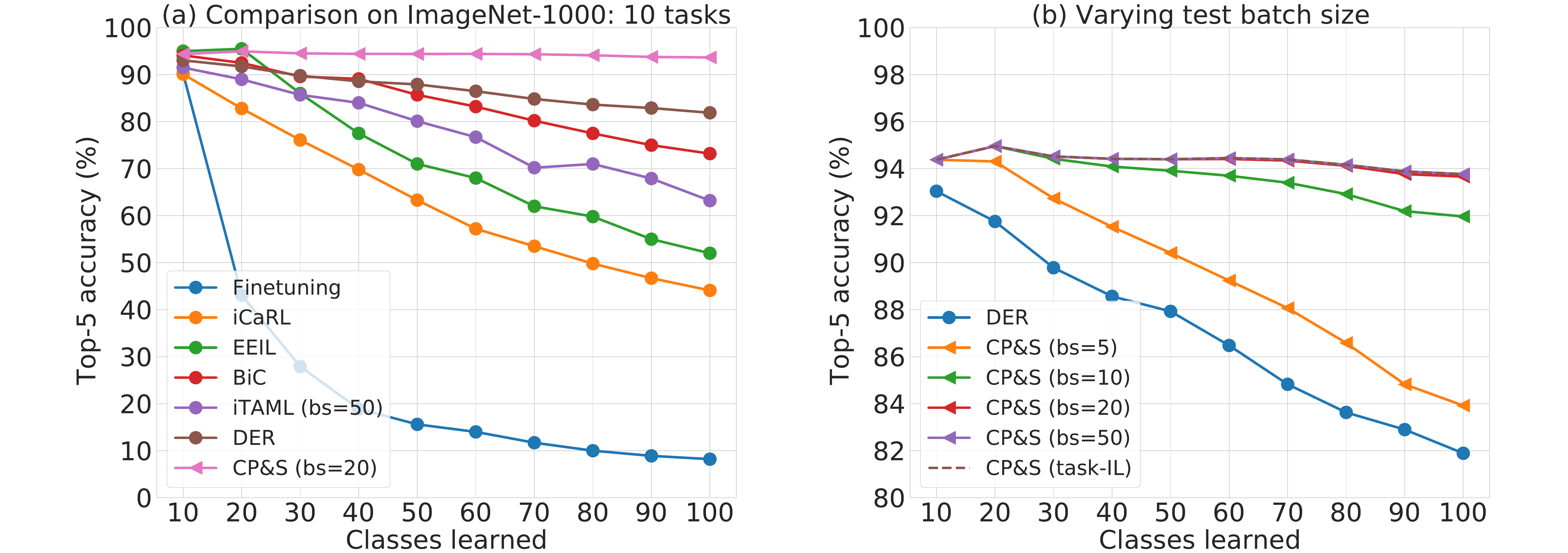}
  \caption{Results obtained for ImageNet-1000 dataset and comparison with other approaches \textbf{(a)} and different batch sizes \textbf{(b)}. Notation: ``bs'' refers to the test batch size; ``task-IL'' refers to the task-IL scenario (upper bound obtained when task ID is known). The pruning parameter is $\alpha_{conv}=0.9$ for CP\&S. Class ordering is generated by seed 1993 (referring to iCaRL's seed).} \label{fig:imagenet1000_resnet18_acc}
\end{figure}

\paragraph{ImageNet-1000.} Focusing now on a more challenging dataset, we split ImageNet-1000 into 10 tasks of 100 classes. To evaluate CP\&S, we train ResNet-18 with 90 epochs and SGD with a learning rate equal to 0.1 dividing it by 10 every 30 epochs. Figure \ref{fig:imagenet1000_resnet18_acc}(a) shows that CP\&S performs better than the state-of-the-art, exhibiting more than 10\% higher Top-5 accuracy than the next best method, which is DER \cite{yan2021dynamically} and more than 20\% improvement over the second best BiC \cite{hou2019learning}. We found this result to be particularly striking, since the prediction accuracy remains around 94\% with virtually no forgetting for the first time in the literature, to the best of our knowledge. Figure \ref{fig:imagenet1000_resnet18_acc}(b) also shows results for different test batch sizes for determining the task-ID and corresponding subnetwork. Once again, a batch size of 20 provides a good trade-off between accuracy and sample size. Interestingly, prediction accuracy is better for CP\&S method than others even when using only 5 test samples in the batch. With 20 images in the test batch, we can almost reach the upper bound of the task-IL scenario, completely reaching it when using 50 images (i.e. identifying the task-ID correctly in $100\%$ of the cases).

In addition, we provide a comparison on the ImageNet-1000 dataset calculating Top-1 accuracy. In Table \ref{tab:aia_imagenet1000}, we observe that CP\&S outperforms the two most recent state-of-the-art methods, DER and FOSTER, by more than 10\%.

\begin{table}
\begin{center}
    \caption{Average incremental accuracy on ImageNet-1000}
    \centering
    \begin{tabular}{c c}
        \toprule
        Method & Top-1 AIA \\
        \midrule
        iCaRL \cite{rebuffi2017icarl} & 38.4\\
        DER \cite{yan2021dynamically} & 66.73 \\
        FOSTER \cite{Wang2022FOSTERFB} & 68.3 \\
        \midrule
        \textbf{CP\&S} (ours) & \textbf{79.08} \\
        \bottomrule
    \end{tabular}
    \label{tab:aia_imagenet1000}
\end{center}    
\end{table}

\paragraph{CUB-200-2011} We split CUB-200-2011 dataset into four tasks with 50 classes in each of the tasks. For testing, we take standard ResNet-18 pretrained on ImageNet-1000 \cite{deng2009imagenet} and fine-tuned with SGD. For iTAML, we also use pretrained weights and use the same hyperparameters for fine-tuning that are used in the original paper for other large-scale datasets.  The pruning parameter for CP\&S is $\alpha_{conv}=0.95$ and only one pruning iteration is used.

\begin{figure}[ht]
  \centering
  \includegraphics[width=\linewidth]{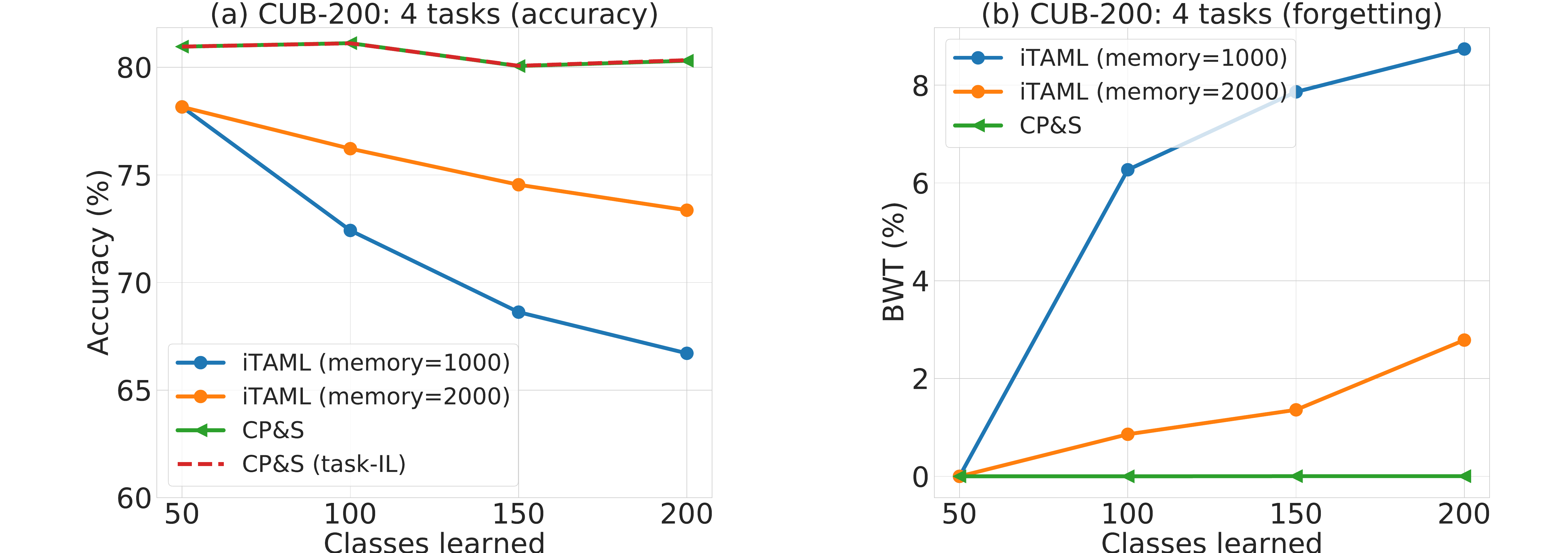}
  \caption{Comparison with iTAML on four tasks constructed from CUB-200-2011. Notation: ``memory'' is the number for images from previous tasks; ``task-IL'' refers to task-IL scenario as an upper-bound for our approach.} \label{fig:cub_200_resnet18_acc_bwt}
\end{figure}

Figure \ref{fig:cub_200_resnet18_acc_bwt} presents the accuracy and BWT history with 20 test images per batch, once again using maxoutput as the task-selection strategy. For 20 test images, it can be observed that CP\&S once again exhibits almost no forgetting of information about previous tasks while learning new ones. However, iTAML even though it keeps 2000 images in memory, continuously forgets previous tasks. In addition, note that 2000 images represent $1/3$ of the CUB-200-2011 dataset, and that we see a dramatic loss of performance for iTAML when using 1000 images (which is still $1/6$ of all the images). In the case of 5 images per test batch, we obtain similar forgetting as iTAML with 2000 images in memory but still a better forgetting metric than iTAML with 1000 images in memory. A more  detailed comparison can be found in Appendix \ref{sec:cub200_appendix}. 

In summary, CP\&S outperformed the state-of-the-art for all datasets considered with the exception of CIFAR-100 when considering a large number of tasks. We demonstrate that we can perform better for small-scale and large-scale datasets (ImageNet-1000) where the second best methods are different. We considered scenarios where each task has a small or a large number of classes, including cases where there is a small number of training examples (CUB-200-2011) without keeping them in memory. 

\section{Further analysis}
\label{sec:further_analysis}

CP\&S method's performance degrades if there are too many tasks because the number of available neuron connections is not enough to create an expressive subnetwork and to select the correct task. This was shown for CIFAR-100 when considering 20 tasks (see Figure \ref{fig:cifar_100_itaml}). In addition, there are scenarios where task selection during inference should be performed by a different strategy instead of maxoutput. For example, when there is an imbalanced number of classes within the tasks we note that using the modification of Importance Scores (IS) to select tasks is advantageous. Focusing on fully connected layers, for the given dataset $\mathbf{X} = \{\mathbf{x}_1, \mathbf{x}_2, \ldots, \mathbf{x}_s\}$ we can compute:
\begin{equation}
    \label{eq:importance_scores}
    \hat{s}^t_{ij} = \overline{w_{ij} \cdot m^t_{ij} \cdot \theta_{i}^{t}(\mathbf{X})} = 
\begin{cases}
    \overline{w_{ij} \cdot \theta_{i}^{t}(\mathbf{X})}, & \text{if there is an active connection } \\ & \text{between neurons $i$ and $j$}\\
    0, & \text{otherwise},\\
\end{cases},
\end{equation}
where $\overline{w_{ij} \cdot \theta_{i}^{t}(\mathbf{X})} = \frac{1}{s}\sum_{k=1}^{s} w_{ij} \cdot \theta_{i}^{t}(\mathbf{x}_k)$ and $\theta_{i}^{t}$ are the feature extractor layers of task $t$.

Suppose we have importance scores $S^1, S^2, \ldots, S^T$ obtained from the training set, we can estimate the importance scores of these connections based on $\mathbf{X}^{test}$ for every subnetwork $t=1,2,\ldots, T$, and denote these estimations as $\hat{S}^1, \hat{S}^2, \ldots, \hat{S}^T$.

Assuming that importance scores should be similar for train and test data for the true task-ID, we can formulate the decisive rule as:
\begin{equation}
    t^{*} = \argmin_{t=1,2,\ldots,T} \sqrt{\sum_{i,j} (s_{ij}^t-\hat{s}_{ij}^t)^2},
\end{equation}
where $s^t_{ij}$ and $\hat{s}^t_{ij}$ are the elements of matrices $S^t$ and $\hat{S}^t$ respectively, $t=1,2,\ldots, T$.

We consider the case where the first task consists of 50 classes, and 10 classes are in each of the following tasks, providing the comparison with iCaRL \cite{rebuffi2017icarl}, LUCIR \cite{hou2019learning}, PODNet \cite{douillard2020podnet} and AFC \cite{kang2022class} (see Table \ref{tab:assumptions} to recall different assumptions for each method). However, we also show that when using IS for task selection we require a larger batch size to improve task identification ($60$ test samples). 

The maxoutput strategy does not work well in this case because most of the parameters are assigned to the first task (with 50 classes). As a result, this strategy predicts the first task when considering the last tasks for almost every batch, as shown in Appendix \ref{sec:cifar100_appendix}. 

\begin{figure}[ht]
  \centering
  \includegraphics[width=\textwidth]{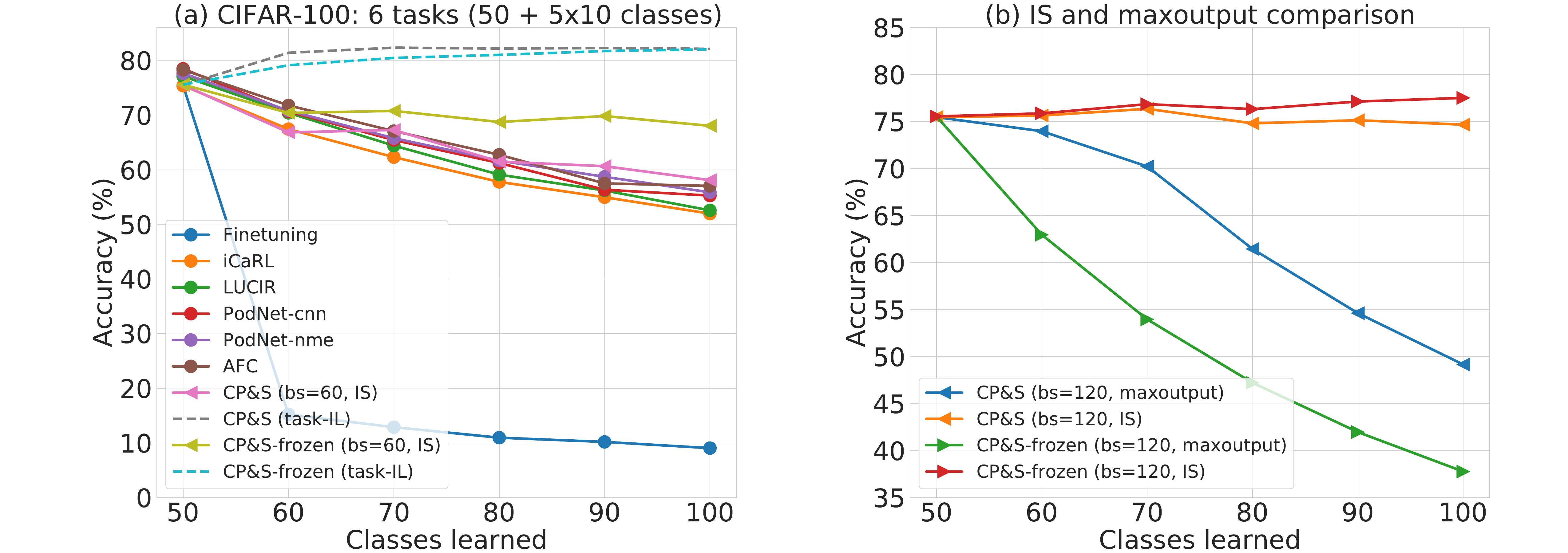}  
  \caption{Accuracy history for ResNet-32 trained with CP\&S and state-of-the-art. The pruning parameter is $\alpha_{conv} = 0.9$ for CP\&S strategy, and ``task-IL'' refers to the same upper bound mentioned in the previous figures. Three different class orderings are used.} \label{fig:resnet32_50+5x10}
\end{figure}

As a final comment, we also investigated an alternative solution when we have a first task that is significantly larger than the following ones. This can be solved by pretraining convolutional weights with the first task and training only the task-specific parts in the network. We denoted this last strategy as ``CP\&S-frozen'' since we pretrain \textit{all} convolutional parameters with the first 50 classes, and, for the next tasks we train task-related batch normalization parameters and the fully connected part that is task-specific by construction. So, in this last strategy each subnetwork consist of a common convolutional part (pretrained on the first task), batch normalization layers and output classification head. We present an additional task-selection accuracy comparison between IS and maxoutput in Appendix \ref{sec:cifar100_appendix}. We again observe poor performance for maximum output response strategy. The final results can be seen in Figure \ref{fig:resnet32_50+5x10} and Table \ref{tab:resnet32_50+5x10}.

\begin{table}[ht]
   \caption{Comparison between algorithms by average incremental accuracy and backward transfer metric at the end of all tasks. Mean values and standard deviation are computed using three different orderings.}
   \centering
    \begin{tabular}{lll}
        Method &  AIA ($\%$) & BWT ($\%$)\\
        \hline
        iCaRL \cite{rebuffi2017icarl} & 61.63 {\scriptsize $\pm$ 0.25} & 12.77 {\scriptsize $\pm$ 0.30}\\
        LUCIR \cite{hou2019learning} & 63.29 {\scriptsize $\pm$ 0.36} & 10.09 {\scriptsize $\pm$ 0.12}\\
        PODNet-CNN \cite{douillard2020podnet} & 64.56 {\scriptsize $\pm$ 0.28} & 11.90 {\scriptsize $\pm$ 0.04} \\
        PODNet-NME \cite{douillard2020podnet} & 65.07 {\scriptsize $\pm$ 0.44} & \textbf{1.18} {\scriptsize $\pm$ 0.16}\\
        AFC \cite{kang2022class} & 65.73 {\scriptsize $\pm$ 0.09} & 7.46 {\scriptsize $\pm$ 0.38} \\
        \hline
        CP\&S (bs=60, IS) & 64.97 {\scriptsize $\pm$ 4.23} & 11.68 {\scriptsize $\pm$ 0.20}\\
        CP\&S-frozen (bs=60, IS) & \textbf{70.55} {\scriptsize $\pm$ 5.05} & 4.90 {\scriptsize $\pm$ 2.35}\\
    \end{tabular}
    \label{tab:resnet32_50+5x10}
\end{table}

We believe this knowledge transfer strategy might be interesting to explore in the future, where the first heads of the network specialize in selecting tasks and the deeper layers specialize in class prediction for each task.  

\begin{figure}[ht!]
  \centering
  \includegraphics[width=\textwidth]{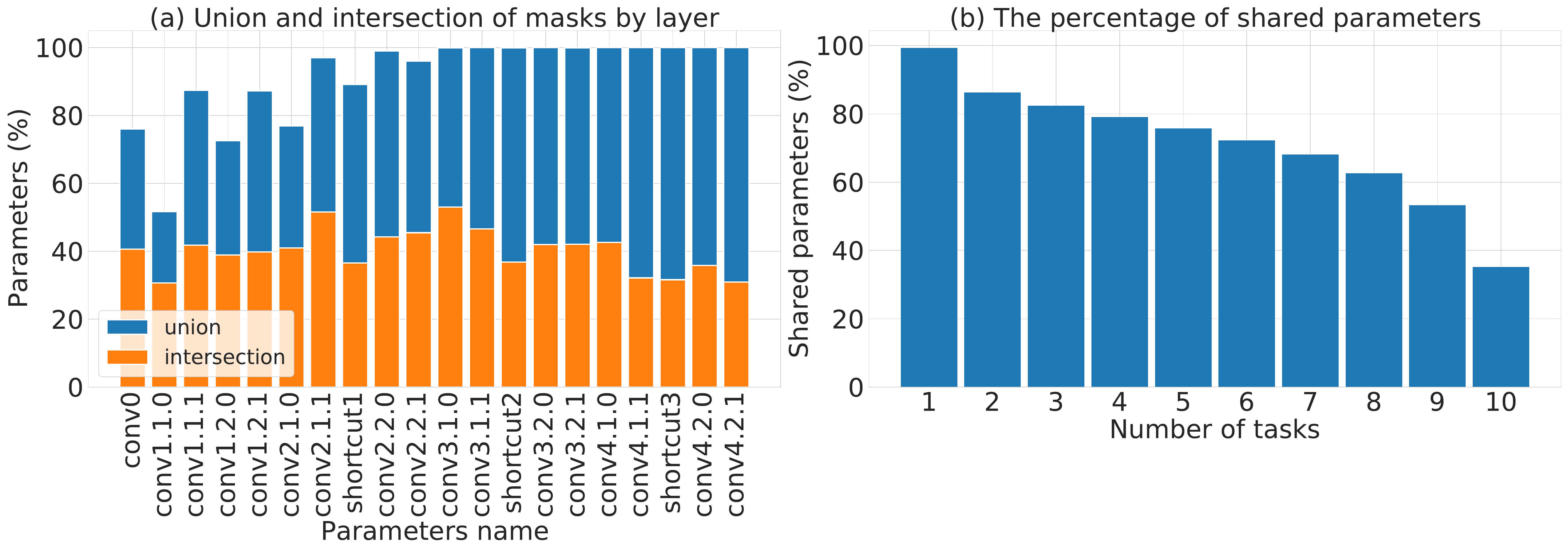}
  \caption{Visualization of employed masks and shared parameters for ResNet-18 on ImageNet-1000.} \label{fig:params_visualization}
\end{figure}

\paragraph{Knowledge transfer}
Let us explore how many parameters are used by every task in the case of ResNet-18 on ImageNet-1000. We consider the union and the intersection of all masks as sets. In Figure \ref{fig:params_visualization}(a) we show how the union and intersection are distributed across parameters after the last task is learned in the case of ResNet-18 on ImageNet-1000. From the union, we observe that the last layers are almost fully occupied in contrast to the first layers. From the intersection, it can be seen that a significant fraction of parameters is shared between each of the tasks across all layers. At the same time, about 85\% are assigned to two and more tasks (Figure \ref{fig:params_visualization}(b)).
Notably, 35\% of parameters are shared across all ten tasks and about 50\% of parameters are used for nine and more tasks. From these figures, we can conclude that almost all parameters are occupied at the end, having significant overlaps between subnetworks. However, looking at Figure \ref{fig:imagenet1000_resnet18_acc}, we see that performance remains stable, without drops. This allows us to conclude that subnetworks share knowledge between tasks, which helps to assimilate new patterns.

\section{Conclusion}
\label{sec:conclusion}

To overcome the problem of catastrophic forgetting while learning new tasks, we propose a continual learning algorithm that trains subnetworks for each task. During training of a task, weights are pruned and then fixed, such that future tasks cannot destroy the weights in this subnetwork, while still being able to use them for other subnetworks. During evaluation of new data, the correct task-ID and associated subnetwork have to be inferred from a small batch of samples. We describe existing task-prediction approaches and propose a new one based on the neural connection strength. Although the task-ID needs to be inferred, no memory is needed to store examples from previous tasks, unlike alternative approaches. The main drawback of the current implementation of the CP\&S strategy is the need to have a small batch of test data due to the difficulty of determining the correct task-ID -- a limitation also observed in the best-performing methods in the literature. Notwithstanding, our work demonstrates that combining subnetwork creation and subnetwork selection methods into one paradigm provides a general approach to solve class-IL problems.  We believe the proposed strategy can be further improved by developing better task-prediction strategies that do not need a batch of test data. CP\&S outperforms all state-of-the-art methods on a variety of datasets. For ImageNet-1000, we show an improvement of more than 10\% accuracy when compared to previous algorithms. Even though we apply CP\&S to image classification tasks, no additional limitations are foreseen when applying it to other machine learning problems. 

\paragraph{Acknowledgement.} The authors would like to thank SURFsara for providing the access to Snellius HPC cluster. 

\bibliographystyle{unsrt}  
\bibliography{references}


\appendix

\section{Appendix}

\subsection{NNrelief details}

\label{appendix:nnrelief}
For the incoming signal $\mathbf{X} = \{\mathbf{x}_1, \mathbf{x}_2, \ldots, \mathbf{x}_N\}$ with $N$ datapoints $\mathbf{x}_n = (x_{n1}, \ldots,  \\ x_{nm_{1}}) \in \mathbb{R}^{m_1}$, NNrelief computes the importance scores: 
\begin{equation}
    s_{ij} (\mathbf{x}_1, \mathbf{x}_2, \ldots, \mathbf{x}_N) = \frac{\overline{|w_{ij} x_{i}|}}{\sum_{k=1}^{m_1} \overline{|w_{kj} x_{k}|}+|b_j|},
\end{equation}

where $\overline{|w_{ij}x_{i}|} = \frac{1}{N}\sum_{n=1}^N |w_{ij}x_{ni}|$ and $\mathbf{W} = (w_{ij}) \in \mathbb{R}^{m_1 \times m_{2}}$ is a corresponding weight matrix, $\mathbf{b} = (b_1, b_2, \ldots, b_{m_2})^{T} \in \mathbb{R}^{m_{2}}$ is a bias vector. Importance score for the bias of the neuron $j$ is $s_{m_1+1,j} = \frac{|b_j|}{\sum_{k=1}^{m_1} \overline{|w_{kj}x_{k}|}+|b_j|}$. 

The sketch of NNrelief algorithm for some fixed neuron $j$ in the following layer is:
\begin{enumerate}
    \item Choose $\alpha \in (0, 1)$ -- the amount of connections' importance that we want to keep relatively to the total importance of the connections coming to the neuron $j$.
    \item Compute importance scores $s_{ij}$ for all connections to the neuron $j, \  i=1, 2, \ldots, m_1+1$.
    \item Sort importance scores $s_{ij}$ for fixed neuron $j$.
    \item For the sorted importance scores $\hat{s}_{ij}$ find minimal $p \le m_1+1$ such that $\sum_{i=1}^p \hat{s}_{ij} \ge \alpha$.
    \item Prune connections with the importance score $s_{ij} < \hat{s}_{pj}$ for all $i \le m_1+1$ and fixed $j$. 
\end{enumerate}

\subsection{Importance scores (IS)}
\label{sec:importance_scores}

Instead of the maxoutput strategy we can consider alternatives to selecting subnetworks. For example, we can use Importance scores (IS). Focusing on fully connected layers and assuming we have importance scores $S^1, S^2, \ldots, S^T$ obtained from the training set, we can estimate the importance scores of these connections based on $\mathbf{X}^{test}$:
\begin{equation}
    \hat{s}_{ij} = \overline{w_{ij} \cdot m^t_{ij} \cdot \theta_{i}^{t}(\mathbf{X}^{test})} = 
\begin{cases}
    \overline{w_{ij} \cdot \theta_{i}^{t}(\mathbf{X}^{test})}, & \text{if there is an active connection } \\ & \text{between neurons $i$ and $j$}\\
    0, & \text{otherwise},\\
\end{cases},
\end{equation}
where $\overline{w_{ij} \cdot \theta_{i}^{t}(\mathbf{X}^{test})} = \frac{1}{s}\sum_{k=1}^{s} w_{ij} \cdot \theta_{i}^{t}(\mathbf{x}_k^{test})$,
and denote these estimations as $\hat{S}^1, \hat{S}^2, \ldots, \hat{S}^T$.
Assuming that importance scores should be similar for train and test data for the true task-ID, we can formulate the decisive rule as:
\begin{equation}
    t^{*} = \argmin_{t=1,2,\ldots,T} \sqrt{\sum_{i,j} (s_{ij}^t-\hat{s}_{ij}^t)^2},
\end{equation}
where $s^t_{ij}$ and $\hat{s}^t_{ij}$ are the elements of matrices $S^t$ and $\hat{S}^t$ respectively, $t=1,2,\ldots, T$.

\subsection{Different task selection strategies on CIFAR-100}
\label{sec:cifar100_appendix}
\begin{figure}[ht!]
  \centering
  \includegraphics[width=\textwidth]{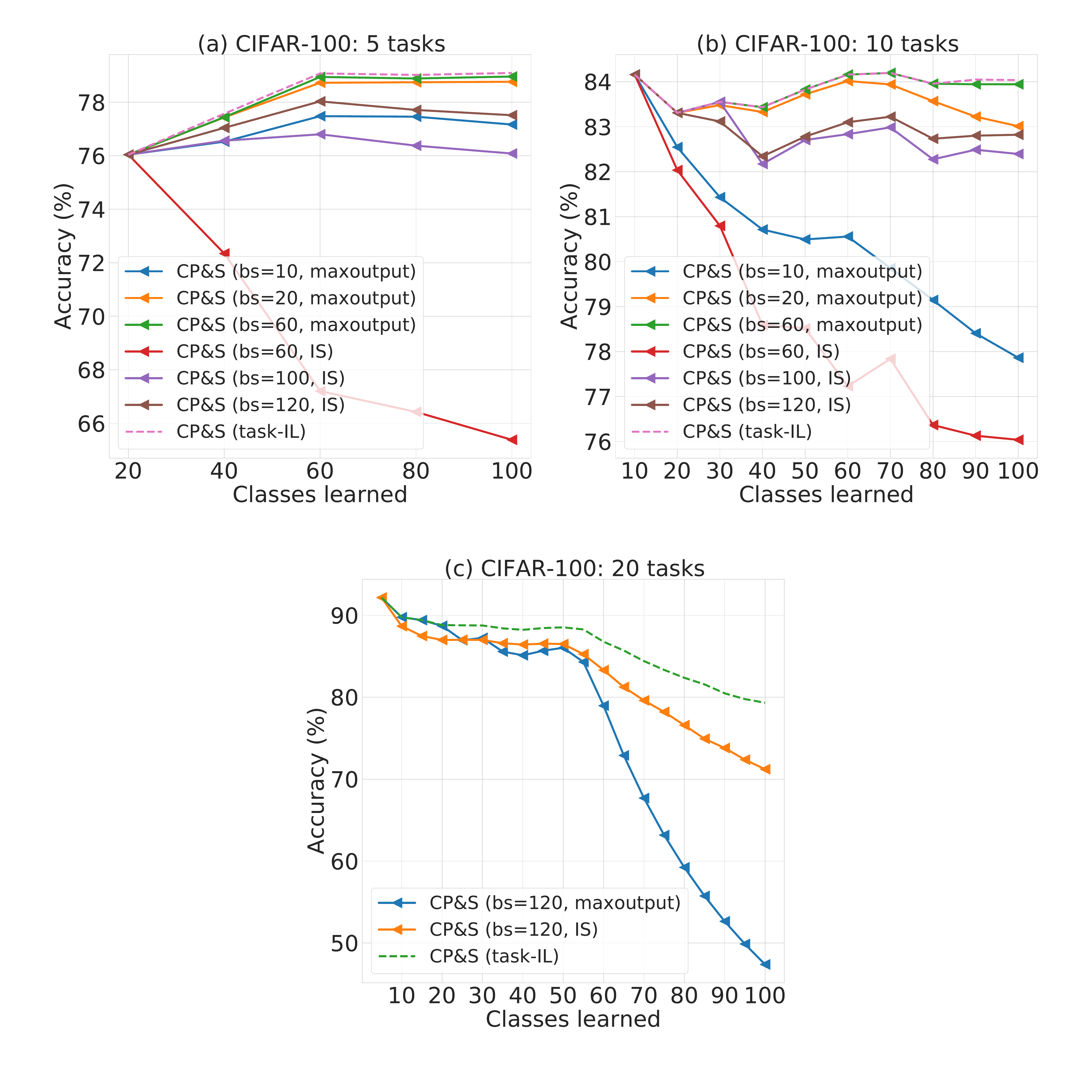}
  \caption{The performance of CP\&S with different batch sizes and task-selection strategies.} \label{fig:cifar_100_ours}
\end{figure}

We present CP\&S results with different test batch sizes and task-selection strategies in Figure \ref{fig:cifar_100_ours}. 

\begin{figure}[ht!]
  \centering
  \includegraphics[width=\linewidth]{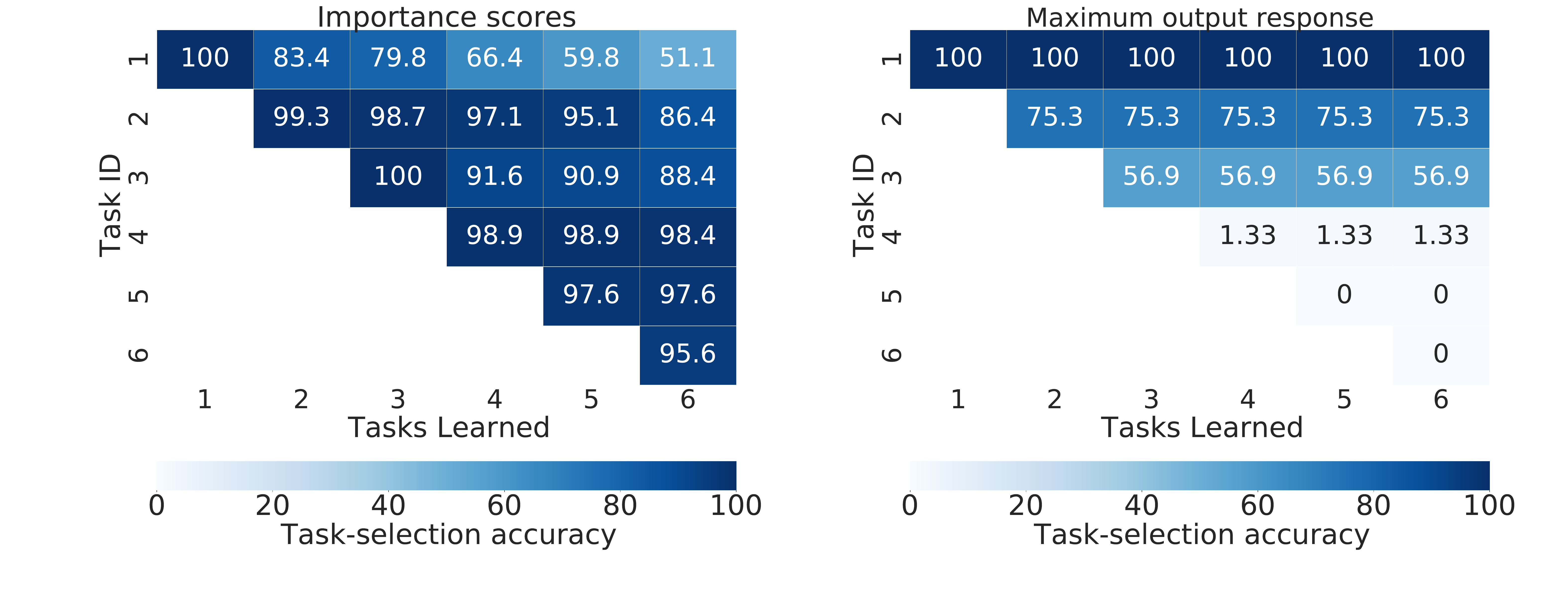}
  \caption{Task-selection accuracy using Importance Scores (IS) (\textbf{left}) as opposed to maxoutput (\textbf{right}) on CIFAR-100 with class imbalance (50 classes in the first task and 10 classes in each of the following five tasks) for CP\&S. The test batch size is 60 images in both cases.}
  \label{fig:cifar_100_resnet32_task-select_0.9_IS_vs_maxoutput}
\end{figure}

\begin{figure}[ht!]
  \includegraphics[width=\linewidth]{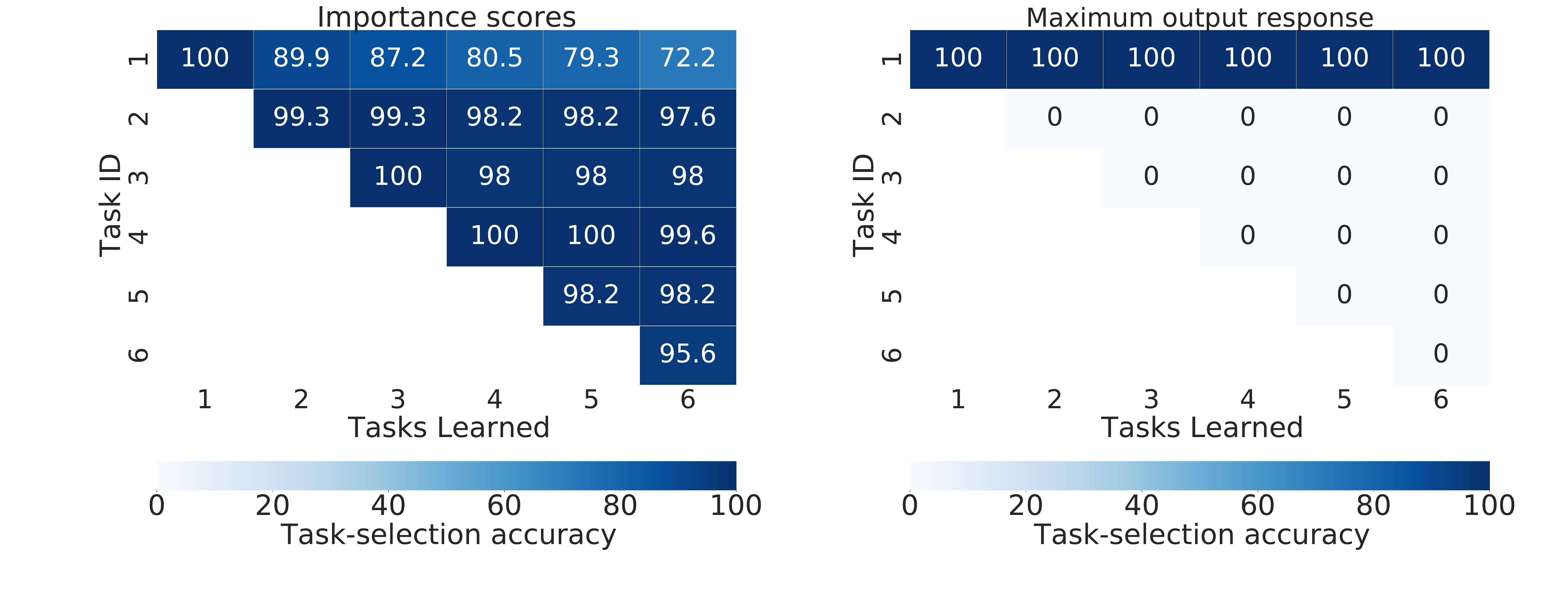}
  \caption{Task-selection accuracy using Importance Scores (IS) (\textbf{left}) as opposed to maxoutput (\textbf{right}) on CIFAR-100 with class imbalance (50 classes in the first task and 10 classes in each of the following five tasks) for CP\&S-frozen. The test batch size is 60 images in both cases.} \label{fig:cifar_100_resnet32_task-select_1_IS_vs_maxoutput}
\end{figure}

Also, we provide an additional comparison between maxoutput and IS strategies in Figs. \ref{fig:cifar_100_resnet32_task-select_0.9_IS_vs_maxoutput} and \ref{fig:cifar_100_resnet32_task-select_1_IS_vs_maxoutput}. In both cases, we observe the advantage of importance scores (IS) over maxoutput strategy in the case of imbalanced tasks.

\subsection{ImageNet-100/1000 results}

For ImageNet-100/1000, we present exact numbers from which the plots are constructed for CP\&S in Tables \ref{tab:imagenet100_results} and \ref{tab:imagenet1000_results}.

\begin{table}[ht!]
\footnotesize
\centering
    \caption{ImageNet-100 results with different test batch sizes and task-IL scenario trained with SGD and Adam.}
    	\begin{tabular}{l p{0.5in} p{0.3in} p{0.3in} p{0.3in} p{0.3in} p{0.3in} p{0.3in} p{0.3in} p{0.3in} p{0.3in} p{0.3in}}
        \toprule
        \bf optimizer & \bf batch size & \bf 1 & \bf 2 & \bf 3 & \bf 4 & \bf 5 & \bf 6 & \bf 7 & \bf 8 & \bf 9 & \bf 10 \\ 
        \midrule
        \multirow{4}{2em}{Adam} & 20  & 98.20 & 98.80 & 98.67 & 98.50 & 98.48 & 98.60 &                                    98.63 & 98.63 & 98.50 & 98.38 \\
                                & 10 & 98.20 & 98.80 & 98.67 & 98.41 & 98.25 & 98.15 &          97.90 & 97.94 & 97.23 & 97.00 \\
                                & 5 & 98.20 & 98.02 & 97.03 & 95.86 & 94.23 & 92.51 & 92.08 & 92.12 & 90.53 & 89.39 \\
                                & task-IL & 98.20 & 98.80 & 98.67 & 98.50 & 98.48 & 98.60 &             98.63 & 98.63 & 98.50 & 98.38 \\
        \midrule
        \multirow{2}{2em}{SGD} & 20 &  99.00 & 98.90 & 98.67 & 98.60 & 97.90 & 98.09 & 98.05 & 98.25 & 94.20 & 92.62 \\
                               & task-IL & 99.00 & 98.90 & 98.67 & 98.60 & 98.20 & 98.33 &             98.26 & 98.43 & 98.20 & 98.06 \\
        \bottomrule
    \end{tabular}
    \label{tab:imagenet100_results}
\end{table}

\begin{table}[ht!]
\centering
\footnotesize
    \caption{ImageNet-1000 results with different test batch sizes and task-IL scenario trained with SGD.}
    \begin{tabular}{l p{0.5in} p{0.3in} p{0.3in} p{0.3in} p{0.3in} p{0.3in} p{0.3in} p{0.3in} p{0.3in} p{0.3in} p{0.3in}}
        \toprule
        \bf optimizer & \bf batch size & \bf 1 & \bf 2 & \bf 3 & \bf 4 & \bf 5 & \bf 6 & \bf 7 & \bf 8 & \bf 9 & \bf 10 \\ 
        \midrule
        \multirow{5}{2em}{SGD} & 50  & 94.38 & 94.96 & 94.52 & 94.42 & 94.40 & 94.45 &                                 94.40 & 94.16 & 93.88 & 93.77 \\
                                & 20 & 94.38 & 94.96 & 94.52 & 94.42 & 94.40 & 94.40 &           94.34 & 94.12 & 93.77 & 93.66 \\
                                & 10 & 94.38 & 94.96 & 94.42 & 94.09 & 93.91 & 93.70 &         93.40 & 92.92 & 92.19 & 91.97 \\
                                & 5 &  94.38 & 94.31 & 92.74 & 91.53 & 90.41 & 89.24 & 88.07 & 86.59 & 84.82 & 83.92 \\
                                & task-IL & 94.38 & 94.96 & 94.52 & 94.42 & 94.40 &                 94.45 & 94.40 & 94.16 & 93.88 & 93.77 \\
        \bottomrule
    \end{tabular}
    \label{tab:imagenet1000_results}
\end{table}

\section{CUB-200-2011 additional comparison}
\label{sec:cub200_appendix}

\begin{figure}[ht!]
  \centering
  \includegraphics[width=\textwidth]{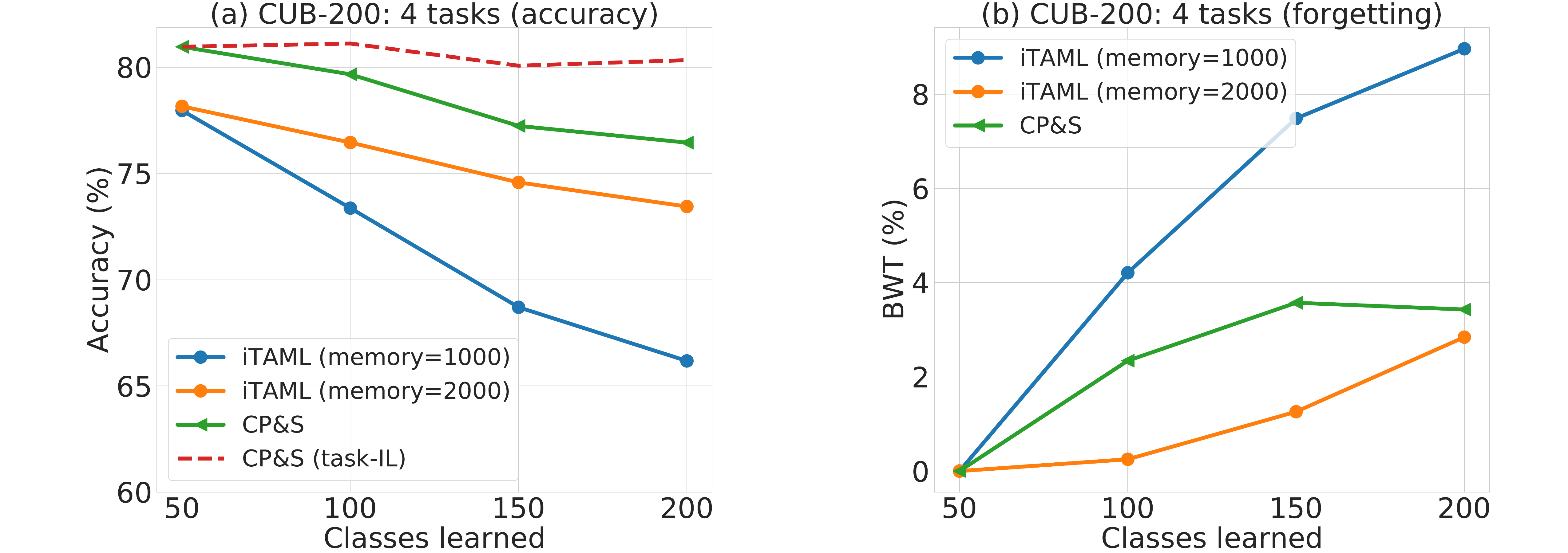}
  \caption{Comparison with iTAML on four tasks constructed from CUB-200-2011. Notation: ``memory'' is the number for images from previous tasks; ``task-IL'' refers to task-IL scenario as an upper-bound for CP\&S.} \label{fig:cub_200_resnet18_acc_bwt_appendix}
\end{figure}

In this section, we provide an additional comparison for ResNet-18 on CUB-200-2011 dataset using 5 test images per batch to predict the task-ID in Fig. \ref{fig:cub_200_resnet18_acc_bwt_appendix}.

\end{document}